\documentclass[10pt,twocolumn,letterpaper]{article}

\usepackage{iccv}
\usepackage{times}
\usepackage{epsfig}
\usepackage{graphicx}
\usepackage{amsmath}
\usepackage{amssymb}

\usepackage{booktabs}
\usepackage{xspace}
\usepackage{multirow}
\usepackage{makecell}
\usepackage{colortbl}
\newcommand{\name}{ReMoDiffuse\xspace}
\usepackage{caption}[font={small}]
\usepackage[dvipsnames]{xcolor}
\usepackage[accsupp]{axessibility}
\usepackage{bbding}
\newcommand\blfootnote[1]{%
  \begingroup
  \renewcommand\thefootnote{}\footnote{#1}%
  \addtocounter{footnote}{-1}%
  \endgroup
}
\newcommand\Mark[1]{\textsuperscript#1}

\usepackage{xcolor}

\usepackage[pagebackref=true,breaklinks=true,letterpaper=true,colorlinks,bookmarks=false]{hyperref}

\iccvfinalcopy 

\def\iccvPaperID{1436} 

\begin{document}

\title{ReMoDiffuse: Retrieval-Augmented Motion Diffusion Model}

\author{Mingyuan Zhang\Mark{1},
Xinying Guo\Mark{1},
Liang Pan\Mark{1},
Zhongang Cai\Mark{1}\Mark{2},
Fangzhou Hong\Mark{1}, 
Huirong Li\Mark{1}, \\
Lei Yang\Mark{2},
Ziwei Liu\Mark{1} \Mark{\Envelope} \\
\Mark{1}S-Lab, Nanyang Technological University, Singapore
\\\Mark{2}Sensetime, China
\\
{\tt\small \{mingyuan001,XGUO012\}@e.ntu.edu.sg, yanglei@sensetime.com, ziwei.liu@ntu.edu.sg}
}

\ificcvfinal\thispagestyle{empty}\fi

\twocolumn[{
    \renewcommand\twocolumn[1][]{#1}
    \maketitle
    \vspace{-10mm}
    \begin{center}
        \includegraphics[width=1.0\linewidth]{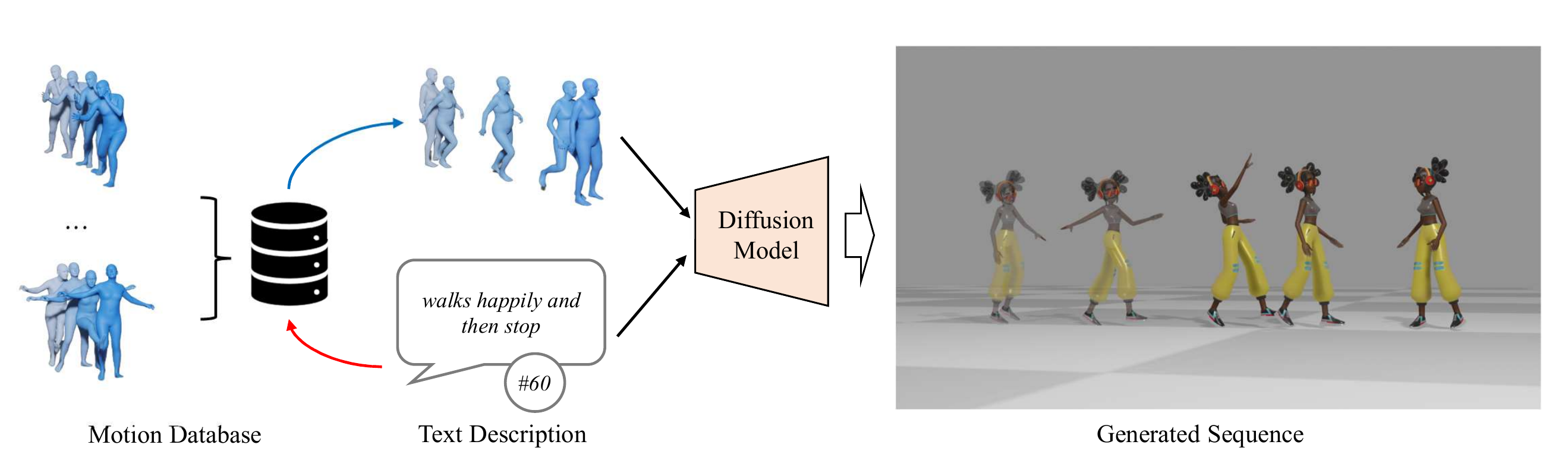}
        \vspace{-10mm}
        \captionof{figure}{\name is a retrieval-augmented 3D human motion diffusion model. Benefiting from the extra knowledge from the retrieved samples, \name is able to achieve high-fidelity on the given prompts.}
        \label{fig:teaser}
    \end{center}
}]

\blfootnote{\Mark{\Envelope} Corresponding author.}
\begin{abstract}
3D human motion generation is crucial for creative industry. Recent advances rely on generative models with domain knowledge for text-driven motion generation, leading to substantial progress in capturing common motions. However, the performance on more diverse motions remains unsatisfactory. 
In this work, we propose \textbf{\name}, a diffusion-model-based motion generation framework that integrates a retrieval mechanism to refine the denoising process.
\name enhances the generalizability and diversity of text-driven motion generation with three key designs:
\textbf{1) Hybrid Retrieval} finds appropriate references from the database in terms of both semantic and kinematic similarities. 
\textbf{2) Semantic-Modulated Transformer} selectively absorbs retrieval knowledge, adapting to the difference between retrieved samples and the target motion sequence. \textbf{3) Condition Mixture} better utilizes the retrieval database during inference, overcoming the scale sensitivity in classifier-free guidance. 
Extensive experiments demonstrate that \name outperforms state-of-the-art methods by balancing both text-motion consistency and motion quality, especially for more diverse motion generation.  Project page: \url{https://mingyuan-zhang.github.io/projects/ReMoDiffuse.html}
\end{abstract}

\section{Introduction}

Human motion generation has numerous practical applications in fields such as game production, film, and virtual reality. This has led to a growing interest in generating manipulable, plausible, diverse, and realistic human motion sequences.  Traditional modeling processes are time-consuming and require specialized equipment and a significant amount of domain knowledge. To address these challenges, generic human motion generation models have been developed to enable the description, generation, and modification of motion sequences. Among all forms of human-computer interaction, natural language, in the form of text, provides rich semantic details and is a commonly used conditional signal in human motion generation.


Previous research has explored various generative models for text-driven motion generation. TEMOS uses a Variational-Auto-Encoder (VAE) to synthesize detailed motions, utilizing the KIT Motion-Language dataset~\cite{plappert2016kit}. Guo \etal \cite{guo2022generating} propose a two-stage auto-regressive approach for generating motion sequences. More recently, diffusion models have been applied to human motion generation due to their strength and flexibility. MotionDiffuse~\cite{zhang2022motiondiffuse} generates realistic and diverse actions while allowing for multi-level motion manipulation in both spatial and temporal dimensions. MDM~\cite{tevet2022human} uses geometric losses as training constraints to make predictions of the sample itself. While these methods have achieved impressive results, they are not versatile enough for uncommon condition signals.

Some recent works on text-to-image generation utilize retrieval methods to complement the model framework, providing an retrieval-augmented pipeline to tackle the above issue~\cite{sheynin2022knn,chen2022re,blattmann2022retrieval}. However, simply transferring these methods into text-driven motion generation fields is impractical due to three new challenges. \textit{Firstly}, the similarity between the target motion sequence and the elements in database is complicated. We need to evaluate both semantic and kinematic similarities to find out related knowledge. \textit{Secondly}, a single motion sequence usually contains several atomic actions. It is necessary to learn from the retrieved samples selectively. In this procedure, the model should be aware of the semantic difference between the given prompt and retrieved samples. \textit{Lastly}, motion diffusion models are sensitive to the scale in classifier-free guidance, especially when we supply another condition, retrieved samples. 

In this paper, we propose a new text-driven motion generation pipeline,  \name, which addresses the abovementioned challenges and thoroughly benefits from the retrieval techniques to generate diverse and high-quality motion sequences. \name includes two stages: retrieval stage and refinement stage. In the retrieval stage, we expect to acquire the most informative samples to provide useful guidance for the denoising process. Here we consider both semantic and kinematic similarities and suggest a \textbf{Hybrid Retrieval} technique to achieve this objective. In the refinement stage, we design a \textbf{Semantics-Modulated Transformer} to leverage knowledge retrieved from an extra multi-modal database and generate semantic-consistent motion sequences. During inference, \textbf{Condition Mixture} technique enables our model to generate high-fidelity and description-consistent motion sequences. We evaluate our proposed \name on two standard text-to-motion generation benchmarks, HumanML3D~\cite{guo2022generating} and KIT-ML~\cite{plappert2016kit}. Extensive quantitative results demonstrate that \name outperforms other existing motion generation pipelines by a significant margin. Additionally,  we propose several new metrics for quantitative comparisons on uncommon samples. We find that \name significantly improves the generation quality on rare samples, demonstrating its superior generalizability.

To summarize, our contributions are threefold: \textbf{1)} We carefully design a retrieval-augmented motion diffusion model which efficiently and effectively explores the knowledge from retrieved samples; \textbf{2)} We suggest new metrics to evaluate the model's generalizability under different scenarios comprehensively; \textbf{3)} Extensive qualitative and quantitative experiments show that our generated motion sequences achieve higher generalizability on both common and uncommon prompts.
\section{Related Work}

\begin{figure*}[t]
    \centering
    \includegraphics[width=1.0\linewidth]{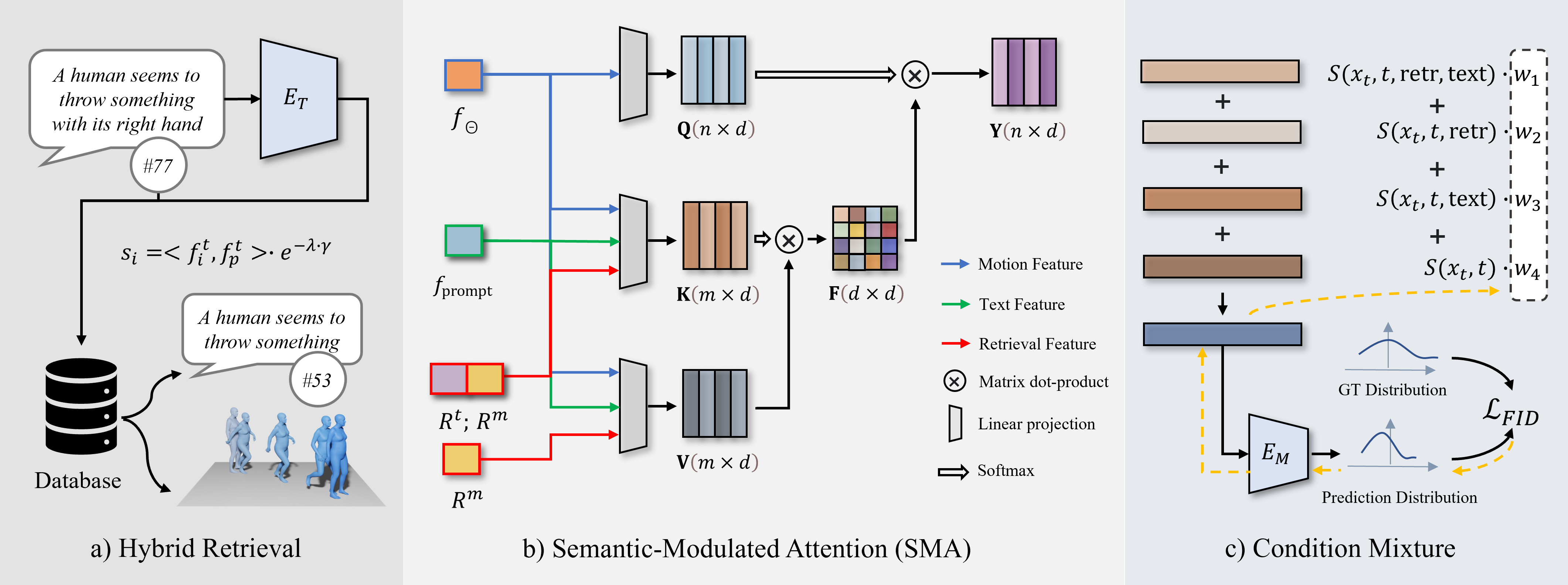}
    \caption{\textbf{Overview} of the proposed \name. a) Hybrid retrieval database stores various features of each training data. The pre-processed text feature and relative difference of motion length are sent to calculate the similarity with the given language description. The most similar ones are fed into the semantics-modulated transformer (SMT), serving as additional clues for motion generation. b) Semantics-modulated transformer incorporates $N$ identical decoder layers, including a semantics-modulated attention (SMA) layer and an FFN layer. The figure shows the detailed architecture of SMA module. CLIP's extracted text features $f_{\mathrm{prompt}}$ from the given prompt, features $R^t$ and $R^m$ from the retrieved samples, and current motion features $f_{\Theta}$ will further refine the noised motion sequence.  c) To synthesize diverse and realistic motion sequences, starting from the pure noised sample, the motion transformer repeatedly eliminates the noise. To better mix outputs under different combinations of conditions, we suggest a training strategy to find the optimal hyper-parameters $w_1,w_2,w_3$ and $w_4$.}
    \label{fig:pipeline}
\end{figure*}

\subsection{Diffusion Models}

Diffusion models~\cite{ho2020denoising, nichol2021improved} is a new class of generative models that have achieved impressive progress on text-to-image generation tasks. Prafulla Dhariwal and Alex Nichol~\cite{dhariwal2021diffusion} propose a diffusion model-based generative model, which first outperforms Generative Adversarial Networks(GAN) and establishes a new state-of-the-art text-driven image generation task. Their success with this advanced generative model quickly attract attention from worldwide researchers. GLIDE~\cite{nichol2021glide} designs classifier-free guidance and proves its superiority compared to the CLIP guidance used in previous works. DALL-E2~\cite{ramesh2022hierarchical} attempts to bridge the text embedding and image embedding from the CLIP~\cite{radford2021learning}. It includes another diffusion model which tries to synthesize an image embedding from the text embedding. 

Recently, some works have focused on employing retrieval methods as complements to the model framework, providing an idea to enhance the generalizability. KNN-Diffusion~\cite{sheynin2022knn} uses k-Nearest-Neighbors (kNN) to train an efficient text-to-image model without any text, enabling the model to adapt to novel samples. RDM~\cite{blattmann2022retrieval} replaces the retrieval examples with the user-assigned images. Then it can effectively transfer artistic style from these images into the generated one. Re-Imagen~\cite{chen2022re} leverages knowledge from the external database to free the model from memorizing rare features, striking a good balance between fidelity and diversity. 

\subsection{Text-Driven Motion Generation}

Text-driven motion generation has witnessed significant progress recently. Earlier works focus on learning a joint embedding space between motion sequences and language descriptions deterministically. JL2P~\cite{ahuja2019language2pose} attempts to create a joint embedding space by applying the same reconstruction task on both text and motion embedding. Specifically, JL2P encodes the input text and motion data separately by two different encoders for each modality. A motion decoder is then applied on both embeddings to reconstruct the original motion sequences, which are expected to be the same as the initial input. Ghosh \etal~\cite{ghosh2021synthesis} further develop this idea by manually dividing each pose sequence into an upper one and a lower one to represent two different body parts. In addition, the proposed method integrates a pose discriminator to improve the generation quality further. MotionCLIP ~\cite{tevet2022motionclip} attempts to enhance the generalizability of text-to-motion generation. It enforces the motion embedding to be similar to the text and image embedding from critical poses. These two embeddings are acquired from CLIP~\cite{radford2021learning}, which excels at encoding texts and images into a joint space. Consequently, MotionCLIP can generate motion sequences with unseen descriptions.

To improve the diversity of generated motion sequences, previous works introduce variational mechanisms. TEMOS ~\cite{petrovich2022temos} employs a Variational Autoencoder (VAE)~\cite{kingma2013auto} to replace the deterministic auto-encoder structures. Besides, different from the recurrent neural networks in the previous works, both motion encoder and motion decoder in TEMOS is based on transformer architectures~\cite{vaswani2017attention}. Guo \etal ~\cite{guo2022generating} propose an auto-regressive conditional VAE, which is conditioned on both the text feature and the previously generated frames. Given these conditions, the proposed pipeline will generate four successive frames as a unit. TEACH~\cite{athanasiou2022teach} also exploits auto-regressive models but in a larger length range. It can synthesize a long motion sequence with the given description and the previous sequence. Consequently, it can generate motion sequences with different actions continuously. TM2T~\cite{guo2022tm2t} regards the text-driven motion generation task as a translation task between natural languages and motion sequences. Most recently, T2M-GPT~\cite{zhang2023generating} quantizes motion clips into discrete tokens and use a transformer to automatically generate later tokens.

Inspired by the success of diffusion models in text-to-image generation tasks, some recent works have adapted this advanced generative model to motion generation tasks. MotionDiffuse~\cite{zhang2022motiondiffuse} is an efficient DDPM-based architecture for plausible and controllable text-driven motion generation. It generates realistic and diverse actions and allows for multi-level motion manipulation in both spatial and temporal dimensions. MDM~\cite{tevet2022human} is a lightweight diffusion model featuring a transformer-encoder backbone. It makes predictions of the sample rather than the noise so that geometric losses are supported as training constraints. Although these methods have outstanding performances on text-driven motion generation tasks, they are not versatile enough for uncommon condition signals. In this paper, we equip the diffusion model-based architecture with retrieval capability, enhancing the generalizability.

\section{Our Approach}
\label{sec:overview}

In this paper, we present a \textbf{Re}trieval-augmented \textbf{Mo}tion \textbf{Diffu}sion model (\textbf{\name}). We first describe the overall architecture of the proposed method in Section ~\ref{sec:overview}. The background knowledge about the motion diffusion model will be discussed in Section ~\ref{sec:mdm}. Then we will introduce our proposed novel retrieval techniques and the corresponding model structure in Section ~\ref{sec:retreival}. Finally, we will introduce the training objective and sampling strategy in Section ~\ref{sec:misc}.

\subsection{Framework Overview}

Figure~\ref{fig:pipeline} shows the overall architecture of \name. We establish the whole pipeline based on MotionDiffuse~\cite{zhang2022motiondiffuse}, which incorporates diffusion models and a series of transformer decoder layers. To strengthen its generalizability, we extract features from two different modalities to establish the retrieval database. During denoising steps, \name first retrieves motion sequences based on the extracted text features and relative motion length. These retrieved samples are then fed into the motion transformer layers. As for each decoder layer, the noised sequence is refined by Semantics-Modulated Attention (SMA) layers and then absorbs information from the given description and the retrieved samples. In the classifier-free generation process, we have distinct outputs under different condition combinations. To better fuse these outputs, we finetune our model on the training split to find the optimal combination of hyper-parameters $w_1, w_2, w_3$ and $w_4$. We will introduce these components in the following subsections. 

\subsection{Diffusion Model for Motion Generation}
\label{sec:mdm}


Recently, diffusion models have been introduced into motion generation~\cite{zhang2022motiondiffuse, tevet2022human}. Compared to VAE-based pipelines, the most popular motion-generative models in previous works, diffusion models strengthen the generation capacity through a stochastic diffusion process, as evidenced by the diverse and high-fidelity generated results. Therefore, in this work, we build our motion generation framework in a corporation with diffusion models.

Diffusion Models can be parameterized as a Markov chain $p_{\theta}(\mathbf{x}_0)\,:=\,\int{p_{\theta}(\mathbf{x}_{0:T})\,d{\mathbf{x}_{1:T}}}$, where $\mathbf{x}_{1},\cdots,\mathbf{x}_{T}$ are the noised sequences distorted from the real data $\mathbf{x}_0 \sim q(\mathbf{x}_0)$. All $\mathbf{x}_t$, where $t=0, 1, 2, \dots, T$, are of the same dimensionality. In the motion generation tasks, each $\mathbf{x}_t$ can be represented by a series of pose $\theta_i \in \mathbb{R}^D, i=1,2,\dots,F$, where $D$ is the dimensionality of the pose representation and $F$ is the number of the frames.

In the forward process of diffusion models, the computation of the posterior distribution $q(\mathbf{x}_{1:T} \vert \mathbf{x}_0)$ is implemented as a Markov chain that gradually adds Gaussian noises to the data according to a variance schedule $\beta_1, \cdots, \beta_T$:
\begin{equation}
    \begin{aligned}
        &q(\mathbf{x}_{1:T} \vert \mathbf{x}_0) \,:=\, \prod_{t=1}^{T} q(\mathbf{x}_t \vert \mathbf{x}_{t-1}), \\
        &q(\mathbf{x}_t \vert \mathbf{x}_{t-1}) \,:=\, \mathcal{N}(\mathbf{x}_t; \sqrt{1-\beta_t}\mathbf{x}_{t-1}, \beta_t\mathbf{I}).
    \end{aligned}
\end{equation}
\par

To efficiently acquire $\mathbf{x}_t$ from $x_0$, Ho \etal~\cite{ho2020denoising} approximate $q(\mathbf{x}_t)$ as $\mathbf{x}:=\sqrt{\bar{\alpha}_t} \mathbf{x}_0 + \sqrt{1 - \bar{\alpha}_t}\epsilon$, where $\alpha_t := 1 - \beta_t$ and $\bar{\alpha}_t:=\prod_{s=1}^t \alpha_s$.

In diffusion models, the aforementioned forwarding Markov chain is reversed to learn the original motion distributions. Expressly, diffusion models are trained to denoise noisy data $\mathbf{x}_t$ into clean data $\mathbf{x}_0$. Following MDM~\cite{tevet2022human}, we predict the clean state $\mathbf{x}_0$. The training target can be written as:
\begin{equation}
\label{eq:objective}
    \mathbb{E}_{x_0,\epsilon,t}[\mathbf{x}_0 - S(\mathbf{x}_t,t,\mathrm{retr},\mathrm{text})],
\end{equation}
where $\mathrm{retr}$ and $\mathrm{text}$ denote the conditions of retrieved samples and the given prompts respectively. Here $t \in \mathcal{U}(0, T)$ denotes the timestamp, which is uniformly sampled from $0$ to the maximum diffusion steps $T$. $S(\mathbf{x}_t,t,\mathrm{retr},\mathrm{text})$ indicates the estimated clean motion sequence, given the four inputs.

During the sampling process, we can sample $\mathbf{x}_{t-1}$ from a Gaussian Distribution $\mathcal{N}(\mu_{\theta}(\mathbf{x}_t,t,c), \beta_t)$, where $c$ denotes the condition of $\mathrm{retr}$ and $\mathrm{text}$ for simplicity. The mean of this distribution can be acquired from $\mathbf{x}_t$ and $S(\mathbf{x}_t,t,c)$ by the following equation:
\begin{equation}
    \begin{aligned}
    & \mu_{\theta}(\mathbf{x}_t,t,c) = \sqrt{\bar{\alpha}_t} S(\mathbf{x}_t,t,c) + \sqrt{1 - \bar{\alpha}_t}\epsilon_{\theta}(\mathbf{x}_t,t,c) \\
    & \epsilon_{\theta}(\mathbf{x}_t,t,c)=(\frac{\mathbf{x}_t}{\sqrt{\bar{\alpha}_t}} - S(\mathbf{x}_t,t,c)) \sqrt{\frac{1}{\bar{\alpha}_t}-1}
    \end{aligned}
\end{equation}
Hence, on the basis of diffusion models, the text-driven motion generation pipeline should be able to predict the start sequence $\mathbf{x}_0$, with the given conditions. In this paper, we propose a retrieval technique to enhance this denoising process. We will introduce how we retrieve motion sequences and how to fuse this information.

\subsection{Retrieval-Augmented Motion Generation}
\label{sec:retreival}

Basically, there are two stages in retrieval-based pipelines. The first stage is to retrieve appropriate samples from the database. The second stage is acquiring knowledge from these retrieved samples to refine the denoising process of diffusion models. We will thoroughly introduce these two steps.

\paragraph{Hybrid Retrieval.} To support this process, we need to extract features for calculating the similarities between the given text description and the entities in the database. Considering that the retrieval procedure is not differentiable, we have to utilize pre-trained models instead of using learnable architectures. An intuitive method is to generate text features on both query text and the data points. Thanks to the pre-trained CLIP~\cite{radford2021learning}, we can easily evaluate the semantic similarities from language descriptions. Formally, for each data point $(\mathrm{text_i}, \Theta_i)$, we first calculate $f^{t}_i=E_{T}(\mathrm{text_i})$ as the text-query feature, where $E_{T}$ is the text encoder in the CLIP model.

Text features usually encourage the retrieval process to select samples with high semantic similarities. 
These features play a significant role in retrieving suitable samples. However, there is another kind of feature that is vital but easily overlooked, the relative magnitude between the expected motion length and that of each entity in the database. Hence, the similarity score $s_i$ between $i$-th data point and the given description $\mathrm{prompt}$ and expected motion length $L$ is defined as below:
\begin{equation}
\begin{aligned}
    &s_i = <f^t_i, f^t_p> \cdot e^{-\lambda \cdot \gamma}, \\
    &f^t_p = E_{t}(\mathrm{prompt}), \gamma = \frac{\Vert l_i - L \Vert}{\max\{l_i, L\}},
\label{eq:score}
\end{aligned}
\end{equation}
where $<\cdot,\cdot>$ denotes cosine similarity between the two given feature vectors, $l_i$ is the length of the motion sequence $\Theta_i$. The similarity score $s_i$ becomes larger when text-query is closer to the prompt feature. When the expected motion length is close to the length of one entity, the corresponding $s_i$ will also increase. This property is significant because the motion sequence with a similar length can provide more informative features for the generation. $\lambda$ is a hyper-parameter to balance the magnitude of these two different similarities. 

To establish the retrieval database, we simply select all the training data as entities. Given the number of retrieved samples $k$, prompt, and motion length $L$, we sort all elements by the score $s_i$ in Equation ~\ref{eq:score}. Then the most $k$ similar ones are selected as the retrieved samples $(\mathrm{text}_i, \Theta_i)$ and fed into the semantics-modulated attention components in the motion transformer. We will illustrate the detailed architecture in the next paragraph.


\paragraph{Network Architecture.} Similar to MotionDiffuse~\cite{zhang2022motiondiffuse} and MDM~\cite{tevet2022human}, we build up our pipeline on the basis of transformer layers as shown in Figure~\ref{fig:pipeline}. In both semantics-modulated attention modules and FFN modules, following MotionDiffuse~\cite{zhang2022motiondiffuse}, we add a stylization block to fuse timestamp $t$ into the motion generation process. First, an embedding vector $\mathbf{e_t}$ is obtained from the timestamp $t$. It should be mentioned that the original design in MotionDiffuse also uses an embedding vector from the given prompt, which is not suitable for classifier-free guidance. Then for each block, a residual shortcut is applied between the input $\mathbf{X} \in \mathbb{R}^{n \times d}$ and the output $\mathbf{Y} \in \mathbb{R}^{n \times d}$, where $n$ is the number of elements and $d$ is the dimensionality.

Two major difficulties should be resolved to better explore knowledge from the retrieved samples. First, in the literature of motion diffusion models~\cite{zhang2022motiondiffuse,tevet2022human}, the resolution of motion sequences is not reduced through the denoising process. The maximum length of one motion sequence is around 200 frames in the HumanML3D~\cite{guo2022generating} dataset, leading to a dramatic computational cost, especially when we expect to retrieve more samples. Hence, efficiency is highly prioritized for the information fusion component. Second, the semantic relation between the retrieved samples and given prompts is complicated. For example, `a person is walking forward' and `a person is walking forward slowly' are highly similar. However, these two prompts will lead to two distinct motion sequences regarding pace and intensity. Therefore, the model should know which motion features can be borrowed, guided by the difference between the language descriptions.

Based on these observations, we design two encoders to extract text features and motion features from the retrieved data, respectively. As for motion features, we expect them to be capable of providing low-level information while retaining the computational cost to an acceptable degree. Therefore, we build up a series of encoder layers, which include alternating Semantics-Modulated Attention(SMA) modules and FFN modules. This motion encoder processes raw motion sequences into usable ones. To reduce the computational cost, we down-sample the sequence into $1/4$ original FPS, which is denoted as $R^m \in \mathbb{R}^{F^{\prime} \cdot k \times D}$, where $F^{\prime}$ is the number of frames after down-sampling and $k$ is the number of retrieved samples. This simple strategy greatly decreases the computation with little information lost. As for the text encoder, the feature $R^t \in \mathbb{R}^{k \times D}$ from the last token is supposed to represent the global semantic information. $R^m$ and $R^t$ constitute the features we needed for the purpose of retrieval-based augmentation.

\paragraph{Semantics-Modulated Attention.} These extracted features will be passed to the cross attention component, as shown in Figure ~\ref{fig:pipeline} . The noised motion sequence forms the query vector $Q \in \mathbb{R}^{F \times D}$. As for the key vector $K$ and the value vector $V$, we consider three sources of data: 1) The motion sequence $f_{\Theta} \in \mathbb{R}^{F \times D}$ itself. As shown in Figure ~\ref{fig:pipeline}, our proposed transformer does not contain a self-attention module. Instead, we combine the function self-attention into the SMA; 2) The text condition $f_{\mathrm{prompt}}$, which semantically describes the expected motion sequence and is extracted as in MotionDiffuse~\cite{zhang2022motiondiffuse}. Specifically, the prompt is first fed into the pre-trained CLIP model to get a feature sequence, which is further processed by two learnable transformer encoder layers; 3) Features $R^m, R^t$ from the retrieved samples. We simply concatenate $f_{\Theta}, f_{\mathrm{prompt}}, R^m$ for value vector $V$ and $f_{\Theta}, f_{\mathrm{prompt}}, [R^m;R^t]$ for key vector $K$. Here $[\cdot;\cdot]$ denotes the concatenation of both terms. This design allows our proposed method to fuse low-level motion information from the retrieved samples and also to fully consider the semantic similarities. The acquired vectors $Q, K, V$ are sent to perform Linear Attention~\cite{shen2021efficient} for efficient computation.

\begin{figure}[]
    \centering
    \includegraphics[width=\linewidth]{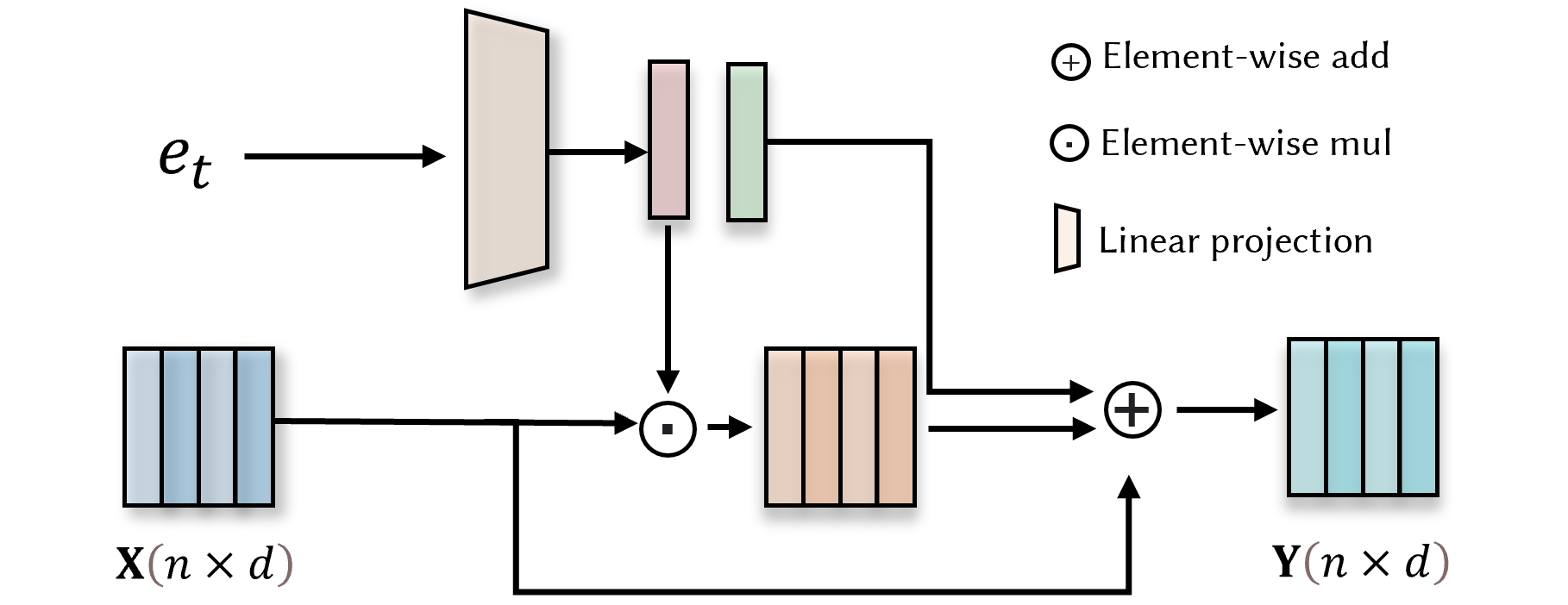}
    \caption{\textbf{Architecture} of the stylization block. This module is adapted from MotionDiffuse~\cite{zhang2022motiondiffuse}. We remove the prompt embedding from the original design to better support classifier-free guidance. This module attempts to inject the information of the current timestamp into the feature representation, which is necessary for denoising steps. Specifically, the timestamp embedding $e_t$ is fed into a series of transformation layers. Two embeddings are generated afterward and serve as an additive offset and a multiplicative offset to the original feature map, respectively. }
    \label{fig:style}
    \vspace{-10pt}
\end{figure}

\paragraph{Stylization Block.} Similar to MotionDiffuse~\cite{zhang2022motiondiffuse} and MDM~\cite{tevet2022human}, we build up our pipeline on the basis of transformer layers. In both semantics-modulated attention modules and FFN modules, following MotionDiffuse~\cite{zhang2022motiondiffuse}, we add a stylization block to fuse timestamp $t$ into the motion generation process. First, an embedding vector $\mathbf{e_t}$ is obtained from the timestamp $t$. Then for each block, a residual shortcut is applied between the input $\mathbf{X} \in \mathbb{R}^{n \times d}$ and the output $\mathbf{Y} \in \mathbb{R}^{n \times d}$, where $n$ is the number of elements and $d$ is the dimensionality. The detailed structure is shown in Figure ~\ref{fig:style}.

\subsection{Condition Mixture}

Classifier-free guidance enables us to generate motion sequences with both high fidelity and consistency with the given text description. A typical formulation is described as below: 
\begin{equation}
\begin{aligned}
\epsilon &= w \cdot \epsilon_{\theta}(\mathbf{x}_t,t,\mathrm{text}) - (w - 1) \cdot \epsilon_{\theta}(\mathbf{x}_t,t),
\end{aligned}
\end{equation}
where $w$ is a hyper-parameter to balance the text-consistency and motion quality. In our proposed retrieval-augmented diffusion pipeline, the given retrieved samples can be regarded as an additional condition. Therefore, we get four estimations: $S(\mathbf{x}_t,t,\mathrm{retr}, \mathrm{text}), S(\mathbf{x}_t,t,\mathrm{retr}), S(\mathbf{x}_t,t,\mathrm{text}), S(\mathbf{x}_t,t)$. We need four parameters to balance these items. To achieve a better performance, here we suggest a \textbf{Condition Mixture} technique to achieve this objective. Specifically, given the pre-trained Semantics-Modulated Transformer (SMT), we optimize the value of $w_1, w_2, w_3, w_4$ and get the final output $\widehat{S}$ as: 
\begin{equation}
\label{eq:output}
\begin{aligned}
\widehat{S} =& w_1 \cdot S(\mathbf{x}_t,t,\mathrm{retr}, \mathrm{text}) + w_2 \cdot S(\mathbf{x}_t,t,\mathrm{text}) + \\
            & w_3 \cdot S(\mathbf{x}_t,t,\mathrm{retr}) + w_4 \cdot S(\mathbf{x}_t,t).
\end{aligned}
\end{equation}
Empirically, we find that the tendency of Frechet Inception Distance (FID) is similar to that of Precision when the hyper-parameters are nearly optimal. Hence, we only attempt to minimize the FID in this procedure. 


\noindent\paragraph{Constrastive Model.} To imitate the evaluator used in the standard evaluation process, we train our contrastive model, which aims at encoding the paired text descriptions and motion sequences into a joint embedding space. As for the motion encoder, we use a 4-layer ACTOR~\cite{petrovich2021action} Encoder. The text encoder is identical to the one we used in \name. The only difference is that we require a sentence feature instead of a sequence of word features. We train this contrastive learning model with the same loss in Guo \etal ~\cite{guo2022generating}. 20K and 40K optimization steps are applied for the KIT-ML and HumanML3D datasets, respectively.

\begin{table*}[ht]
\centering
\caption{\textbf{Evaluation results of different evaluator.}}
\label{tab:evaluator}
\setlength{\tabcolsep}{1.4mm}
{
\begin{tabular}{lcccccccc}
\hline
\multirow{2}{2cm}{\centering Methods} & \multirow{2}{2cm}{\centering Dataset} & \multicolumn{3}{c}{\centering R Precision$\uparrow$} & \multirow{2}{1.5cm}{\centering FID$\downarrow$} & \multirow{2}{2.5cm}{\centering MM Dist$\downarrow$} & \multirow{2}{2cm}{\centering Diversity$\uparrow$} \\
& &  Top 1 & Top 2 & Top 3 \\
\hline
Guo \etal & HumanML3D & $0.511^{\pm .003}$ & $0.703^{\pm.003}$ & $0.797^{\pm.002}$ & $0.002^{\pm.000}$ & $2.974^{\pm.008}$ & $9.503^{\pm.065}$ \\
Ours & HumanML3D & $0.539^{\pm .004}$ & $0.721^{\pm.003}$ & $0.810^{\pm.003}$ & $0.001^{\pm.000}$ & $1.462^{\pm.006}$ & $5.298^{\pm.047}$ \\
\hline
Guo \etal & KIT-ML & $0.424^{\pm .005}$ & $0.649^{\pm.006}$ & $0.779^{\pm.006}$ & $0.031^{\pm.004}$ & $2.788^{\pm.012}$ & $11.08^{\pm.097}$ \\ 
Ours & KIT-ML & $0.475^{\pm .006}$ & $0.690^{\pm.004}$ & $0.791^{\pm.005}$ & $0.002^{\pm.000}$ & $1.337^{\pm.012}$ & $6.371^{\pm.058}$ \\
\hline
\end{tabular}}
\end{table*}

\noindent\paragraph{Parameter Finetuning.} As mentioned before, we only use 50 denoising steps to generate motion sequences in the inference stage. However, it is impractical to calculate the gradient through such several forward times. To simplify the problem, we divide all denoising steps into the first 40 steps and the last ten steps. In the first part, we use grid search to find a better parameter combination. Specifically, for Equation ~\ref{eq:output}, we search $w_1$ and $w_2$ from $[-5, 5]$ with step $0.5$ to find the best parameter for each model. Here we use inspiration from Re-Imagen~\cite{chen2022re} that set $w_4=0$. Besides, to retain the output's statistics, we $w_1 + w_2 + w_3 + w_4=1$. These two properties enable us to find the optimal combination by only searching the value of $w_1$ and $w_2$. The evaluation metric is the calculated FID between our generated sequences and the natural motion sequences in the training split performed by our trained contrastive model. This search aims to find an optimal combination of $w_1$ and $w_2$ to achieve the lowest FID.

In the second stage, we use an end-to-end training scheme to optimize $w_1, w_2$, and $w_3$. $w_4$ is acquired by $1-w_1-w_2-w_3$. We use the Adam optimizer to train our model on the training split for 1K steps to find the best parameter combination.

We use the searched parameters during training to perform the first 40 denoising steps. After that, we auto-regressively denoise the motion sequence with learnable $w_1, w_2$ and $w_3$. The training objective here is also reducing FID.

We use the Adam optimizer and train 1K steps for both HumanML3D and KIT-ML datasets to find the best parameter combination.

\begin{table*}[ht]
\small
\centering
\caption{\textbf{Quantitative results on the HumanML3D test set.} For a fair comparison, all methods use the real motion length from the ground truth as the extra given information. `$\uparrow$'(`$\downarrow$') indicates that the values are better if the metric is larger (smaller). We run all the evaluations 20 times. $x^{\pm y}$ indicates that the average metric is $x$ and the the 95\% confidence interval is $y$. The best result and the second best result are in red cells and blue cells, respectively.}
\label{tab:humanml3d}
\setlength{\tabcolsep}{1.4mm}
{
\begin{tabular}{lccccccc}
\hline

\multirow{2}{2cm}{\centering Methods} & \multicolumn{3}{c}{\centering R Precision$\uparrow$} & \multirow{2}{1.5cm}{\centering FID$\downarrow$} & \multirow{2}{2.5cm}{\centering MM Dist$\downarrow$} & \multirow{2}{2cm}{\centering Diversity$\uparrow$} & \multirow{2}{2cm}{\centering MultiModality$\uparrow$} \\
& Top 1 & Top 2 & Top 3 \\
\hline
Real motions & $0.511^{\pm .003}$ & $0.703^{\pm.003}$ & $0.797^{\pm.002}$ & $0.002^{\pm.000}$ & $2.974^{\pm.008}$ & $9.503^{\pm.065}$ & -\\ 
\hline

Language2Pose~\cite{ahuja2019language2pose} & $0.246^{\pm.002}$ & $0.387^{\pm.002}$ & $0.486^{\pm.002}$ & $11.02^{\pm.046}$ & $5.296^{\pm.008}$ & $7.676^{\pm.058}$ & - \\

Text2Gesture~\cite{bhattacharya2021text2gestures} & $0.165^{\pm.001}$ & $0.267^{\pm.002}$ & $0.345^{\pm.002}$ & $7.664^{\pm.030}$ & $6.030^{\pm.008}$ & $6.409^{\pm.071}$ & - \\

MoCoGAN~\cite{tulyakov2018mocogan} & $0.037^{\pm.000}$ & $0.072^{\pm.001}$ & $0.106^{\pm.001}$ & $94.41^{\pm.021}$ & $9.643^{\pm.006}$ & $0.462^{\pm.008}$ & $0.019^{\pm.000}$ \\

Dance2Music~\cite{lee2019dancing} & $0.033^{\pm.000}$ & $0.065^{\pm.001}$ & $0.097^{\pm.001}$ & $66.98^{\pm.016}$ & $8.116^{\pm.006}$ & $0.725^{\pm.011}$ & $0.043^{\pm.001}$ \\

Guo \etal~\cite{guo2022generating}  & $0.457^{\pm.002}$ & $0.639^{\pm.003}$ & $0.740^{\pm.003}$ & $1.067^{\pm.002}$ & $3.340^{\pm.008}$ & $9.188^{\pm.002}$ & \cellcolor{blue!25}$2.090^{\pm.083}$ \\

\hline

MDM~\cite{tevet2022human} & - & - & $0.611^{\pm.007}$ &  $0.544^{\pm.044}$ & $5.566^{\pm.027}$ & $9.559^{\pm.086}$ & \cellcolor{red!25}$2.799^{\pm.072}$ \\

MotionDiffuse~\cite{zhang2022motiondiffuse} & \cellcolor{blue!25} $0.491^{\pm.001}$ & \cellcolor{blue!25} $0.681^{\pm.001}$ & \cellcolor{blue!25} $0.782^{\pm.001}$ & $0.630^{\pm.001}$ & \cellcolor{blue!25} $3.113^{\pm.001}$ & \cellcolor{red!25} $9.410^{\pm.049}$ &  $1.553^{\pm.042}$ \\

T2M-GPT~\cite{zhang2023generating} & \cellcolor{blue!25} $0.491^{\pm.003}$ & $0.680^{\pm.003}$ & $0.775^{\pm.002}$ & \cellcolor{blue!25} $0.116^{\pm.004}$ & $3.118^{\pm.011}$ & \cellcolor{blue!25} $9.761^{\pm.081}$ & $1.856^{\pm .011}$ \\

\hline
Ours & \cellcolor{red!25} $0.510^{\pm.005}$ & \cellcolor{red!25} $0.698^{\pm.006}$ & \cellcolor{red!25} $0.795^{\pm.004}$ & \cellcolor{red!25} $0.103^{\pm.004}$ & \cellcolor{red!25} $2.974^{\pm.016}$ & $9.018^{\pm.075}$ & $1.795^{\pm.043}$ \\
\hline
\vspace{-10pt}
\end{tabular}}
\end{table*}

\begin{table*}[ht]
\centering
\small
\caption{\textbf{Quantitative results on the KIT-ML test set.}}
\label{tab:kit}
\vspace{-10pt}
\setlength{\tabcolsep}{1.4mm}
{
\begin{tabular}{lccccccc}
\hline

\multirow{2}{2cm}{\centering Methods} & \multicolumn{3}{c}{\centering R Precision$\uparrow$} & \multirow{2}{1.5cm}{\centering FID$\downarrow$} & \multirow{2}{2.5cm}{\centering MM Dist$\downarrow$} & \multirow{2}{2cm}{\centering Diversity$\uparrow$} & \multirow{2}{2cm}{\centering MultiModality$\uparrow$} \\
& Top 1 & Top 2 & Top 3 \\
\hline
Real motions & $0.424^{\pm .005}$ & $0.649^{\pm.006}$ & $0.779^{\pm.006}$ & $0.031^{\pm.004}$ & $2.788^{\pm.012}$ & $11.08^{\pm.097}$ & -\\ 
\hline

Language2Pose~\cite{ahuja2019language2pose} & $0.221^{\pm.005}$ & $0.373^{\pm.004}$ & $0.483^{\pm.005}$ & $6.545^{\pm.072}$ & $5.147^{\pm.030}$ & $9.073^{\pm.100}$ & - \\

Text2Gesture~\cite{bhattacharya2021text2gestures} & $0.156^{\pm.004}$ & $0.255^{\pm.004}$ & $0.338^{\pm.005}$ & $12.12^{\pm.183}$ & $6.964^{\pm.029}$ & $9.334^{\pm.079}$ & - \\

MoCoGAN~\cite{tulyakov2018mocogan} & $0.022^{\pm.002}$ & $0.042^{\pm.003}$ & $0.063^{\pm.003}$ & $82.69^{\pm.242}$ & $10.47^{\pm.012}$ & $3.091^{\pm.043}$ & $0.250^{\pm.009}$ \\

Dance2Music~\cite{lee2019dancing} & $0.031^{\pm.002}$ & $0.058^{\pm.002}$ & $0.086^{\pm.003}$ & $115.4^{\pm.240}$ & $10.40^{\pm.016}$ & $0.241^{\pm.004}$ & $0.062^{\pm.002}$ \\

Guo \etal~\cite{guo2022generating}  & $0.370^{\pm.005}$ & $0.569^{\pm.007}$ & $0.693^{\pm.007}$ & $2.770^{\pm.109}$ & $3.401^{\pm.008}$ & $10.91^{\pm.119}$ & $1.482^{\pm.065}$ \\

\hline


MDM~\cite{tevet2022human}& -  & - & $0.396^{\pm.004}$ & \cellcolor{blue!25}$0.497^{\pm.021}$ & $9.191^{\pm.022}$ & $10.847^{\pm.109}$ & \cellcolor{red!25}$1.907^{\pm.214}$ \\

MotionDiffuse~\cite{zhang2022motiondiffuse} & \cellcolor{blue!25}$0.417^{\pm.004}$ & $0.621^{\pm.004}$ & $0.739^{\pm.004}$ &  $1.954^{\pm.062}$ &  \cellcolor{blue!25}$2.958^{\pm.005}$ & \cellcolor{red!25} $11.10^{\pm.143}$ & $0.730^{\pm.013}$\\

T2M-GPT~\cite{zhang2023generating} &  $0.416^{\pm.006}$ & \cellcolor{blue!25} $0.627^{\pm.006}$ & \cellcolor{blue!25}$0.745^{\pm.006}$ & $0.514^{\pm.029}$ & $3.007^{\pm.023}$ & \cellcolor{blue!25} $10.921^{\pm.108}$ & \cellcolor{blue!25} $1.570^{\pm .039}$ \\
\hline

Ours & \cellcolor{red!25}$0.427^{\pm.014}$ & \cellcolor{red!25}$0.641^{\pm.004}$ & \cellcolor{red!25}$0.765^{\pm.055}$ & \cellcolor{red!25}$0.155^{\pm.006}$ & \cellcolor{red!25}$2.814^{\pm.012}$ & $10.80^{\pm.105}$ & $1.239^{\pm.028}$ \\
\hline
\end{tabular}}
\end{table*}

\subsection{Training and Inference}
\label{sec:misc}
\paragraph{Model Training.} 
Inspired by the classifier-free technique, 10\% of the text conditions and 10\% of the retrieval conditions are independently randomly masked to approximate $p({\mathbf{x}_0})$. The training object is to minimize the mean square error between the predicted initial sequence and the ground truth, as shown in Equation ~\ref{eq:objective}. In the training stage, we typically use a 1000-steps diffusion process.


\paragraph{Model Inference.} 

During each denoising step, we use the learned coefficients $w_1,w_2,w_3$ and $w_4$ to get $\widehat{S}$ as Equation ~\ref{eq:output}. To reduce the computation cost introduced by the retrieved samples, we pre-process all $f^v_i, f^t_i, R^t, R^m$ to ensure no repeated computation for different syntheses.

Different from the training stage, we carefully reduce the whole denoising process into 50 steps during inference, which enables our model to generate high-quality motion sequences efficiently.

\section{Experiments}
\begin{figure*}[ht]
    \centering
    \includegraphics[width=0.95\linewidth]{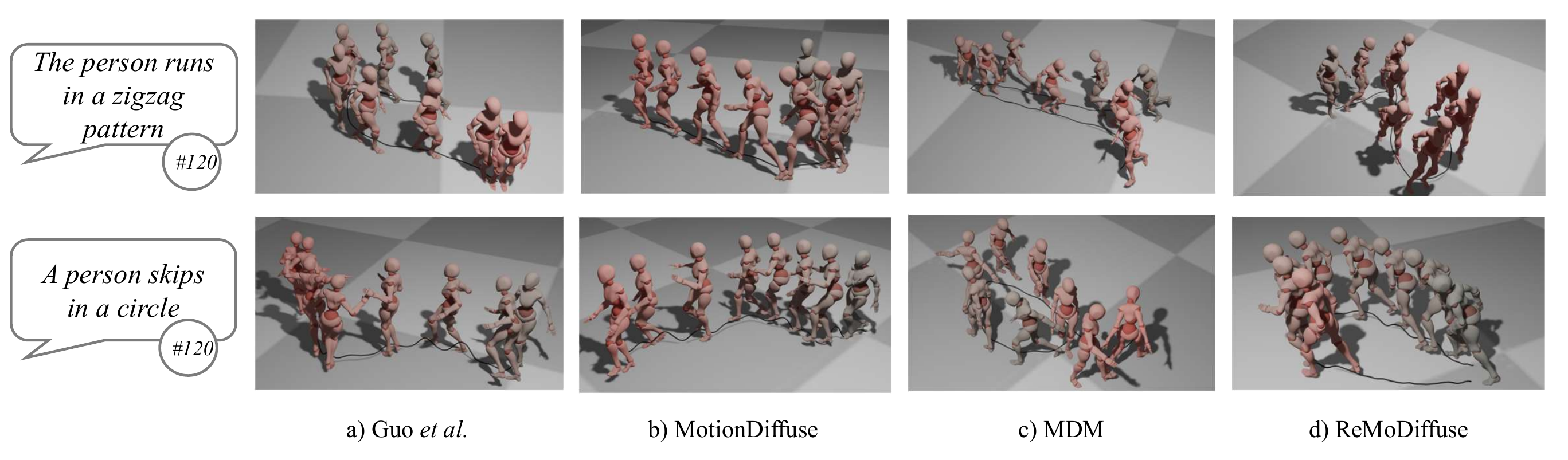}
    \caption{Visual Comparison between previous works and \name. We draw black lines to show the translation path. As for both given conditions, only \name conveys accurate action and path condition.}
    \label{fig:comparison}
\end{figure*}

\subsection{Datasets and Metrics}
\paragraph{Datasets.} We evaluate our proposed framework using the KIT dataset~\cite{plappert2016kit} and the HumanML3D dataset~\cite{guo2022generating}, two leading benchmarks in text-driven motion generation tasks. KIT Motion Language Dataset is an open dataset combining human motion and natural language, which contains 3,911 motions and 6,363 natural language annotations. HumanML3D is a scripted 3D human motion dataset that originates from and textually reannotates the HumanAct12~\cite{guo2020action2motion} and AMASS datasets~\cite{mahmood2019amass}. Overall, HumanML3D consists of 14,616 motions and 44,970 descriptions.
\paragraph{Evaluation Metrics.} We follow the performance measures employed in MotionDiffuse for quantitative evaluations, namely Frechet Inception Distance (FID), R Precision, Diversity, Multimodality, and Multi-Modal Distance. (1) FID is an objective metric calculating the distance between features extracted from real and generated motion sequences, which highly reflects the generation quality. (2) R-precision measures the similarity between the text description and the generated motion sequence and indicates the probability that the real text appears in the top k after sorting, and in this work, k is taken to be 1, 2, and 3. (3) Diversity measures the variability and richness of the generated action sequences. (4) Multimodality measures the average variance of generated motion sequences given a single text description. (5) Multi-modal distance (MM Dist for short) represents the average Euclidean distance between the motion feature and its corresponding text description feature.

\subsection{Implementation Details}

We use similar settings on HumanML3D and KIT-ML datasets. As for the motion encoder, a 4-layer transformer is used, and the latent dimension is 512. As for the text encoder, a frozen text encoder used in the CLIP ViT-B/32, together with 2 additional transformer encoder layers, is built and applied. As for the diffusion model, the variances $\beta_t$ are pre-defined to spread linearly from 0.0001 to 0.02, and the total number of noising steps is set to be $T = 1000$. Adam is adapted as the optimizer to train the model with a learning rate equal to 0.0002. 1 Tesla V100 is used for training, and the batch size on a single GPU is 128. Pieces of training on KIT-ML and HumanML3D are carried out for 40k and 200k steps respectively. \par
Pose representation in this work follows the schema used by Guo \etal~\cite{guo2022generating}. The pose is defined as a tuple of length seven: $(r^{va},r^{vx}, r^{vz},r^h, \mathbf{j}^p, \mathbf{j}^v, \mathbf{j}^r)$, where $r^{va}\in \mathbb{R}$ is the root angular velocity along Y-axis, and $r^{vx}, r^{vz}\in \mathbb{R}$ are the root linear velocities along X-axis and Z-axis respectively. $r^h \in \mathbb{R}$ is the root height. $\mathbf{j}^p, \mathbf{j}^v \in \mathbb{R}^{J \times 3}$ are the local joints positions and velocities. $\mathbf{j}^r \in \mathbb{R}^{J \times 6}$ is the 6D local continuous joints rotations. $J$ denotes the number of joints, and in HumanML3D and KIT-ML, $J$ is 22 and 21 separately.

\subsection{Main Results}


Table ~\ref{tab:humanml3d} and Table ~\ref{tab:kit} show the comparison between our proposed \name and four other existing works, including recent diffusion models-based algorithms~\cite{tevet2022human, zhang2022motiondiffuse}, one VAE-based generative model~\cite{guo2022generating}, and one GPT-style generative model~\cite{zhang2023generating}.

Compared to other diffusion model-based pipelines, our proposed \name achieves a better balance between the condition-consistency and fidelity. It should be noted that, \name is the first work to achieve state-of-the-art on both metrics, which demonstrates the superiority of the proposed pipeline.

\subsection{Ablation Study}

\begin{figure}[t]
    \centering
    \includegraphics[width=1.0\linewidth]{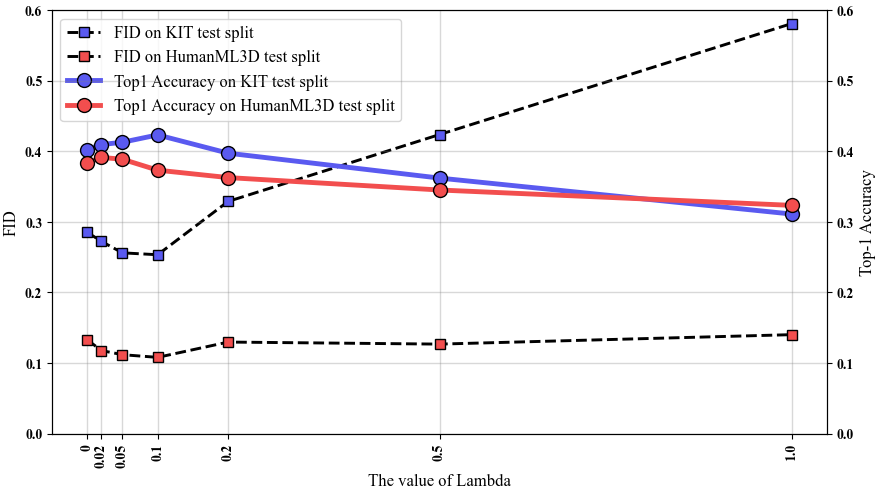}
    \caption{The retrieval performance of different $\lambda$. $\lambda$ is used to balance semantic and kinematic similarity in the retrieval stage. A larger $\lambda$ indicates the retrieval process focuses more on the kinematic similarity.}
    \label{fig:retrieval}
    \vspace{-15pt}
\end{figure}

\paragraph{Retrieval Techniques.} First, we investigate the influence of different retrieval techniques. To directly evaluate the similarity between the target samples and the given samples, we use retrieved samples as generated results and calculate the FID metric for them. We try different $\lambda$ to balance the terms of semantic similarity and kinematic similarity. The results are shown in Figure ~\ref{fig:retrieval}. $\lambda=0$ means that the kinematic similarity will not influence the retrieval process, whose retrieval quality is unacceptable. This result supports our claim that kinematic similarity is significant to the retrieval quality. The optimal value of $\lambda$ is $0.1$ for both KIT-ML and HumanML3D datasets.

\begin{figure*}[h]
    \centering
    \includegraphics[width=\linewidth]{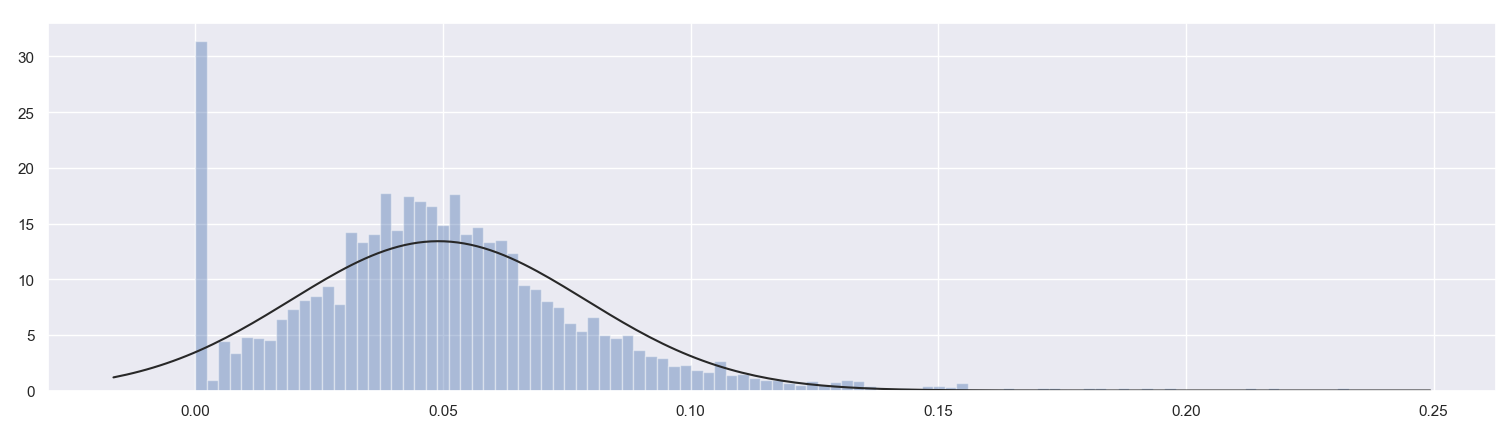}
    \caption{\textbf{Rareness distribution of HumanML3D test split.} We split all testcases into 100 bins according to its Rareness value.}
    \label{fig:dist}
\end{figure*}

\paragraph{Motion Refinement.} We further evaluate the proposed cross attention component of our retrieval-augmented motion generation. In Table ~\ref{tab:cross_attention}, when using the text feature, FID is enhanced remarkably. It strongly supports our claims that text features are highly significant in hybrid retrieval, which is not discussed in the text-to-image generation tasks. Besides, the proposed retrieval techniques outperform the baseline by a remarkable margin.

\begin{table}[t]
\centering
\small
\caption{\textbf{Ablation of the proposed architecture.} All results are reported on the KIT testset. `T' and `M' denote the usage of semantic similarity and kinematic similarity respectively. These two factors are considered in both retrieval and refinement stages.}
\label{tab:cross_attention}
\setlength{\tabcolsep}{1.4mm}
{
\begin{tabular}{cccccc}
\hline

& Retrieval & Attention & \#Samples & Stride & FID$\downarrow$ \\
\hline
a) & - & - & - & - & $0.245^{\pm.008}$ \\
\hline
b) & T & M & 2 & 4 & $0.314^{\pm.012}$ \\
c) & T\& M & M & 2 & 4 & $0.192^{\pm.008}$ \\
d) & T & T \& M & 2 & 4 & $0.307^{\pm.010}$ \\
e) & T\& M & T\& M & 2 & 4 & $0.155^{\pm.006}$ \\
\hline
f) & T\& M & T\& M & 1 & 4 & $0.186^{\pm.008}$ \\
g) & T\& M & T\& M & 3 & 4 & $0.217^{\pm.009}$ \\
\hline
\end{tabular}}
\vspace{-10pt}
\end{table}

\begin{table*}[h]
\centering
\caption{\textbf{Examples of Rareness in the HumanML3D test set.}}
\label{tab:example}
\renewcommand{\arraystretch}{1.4}
\setlength{\tabcolsep}{1.4mm}
{
\begin{tabular}{ccp{12cm}}
\hline

\textbf{Rareness} & \textbf{Rareness Quantile} & \textbf{Caption} \\
\hline
$0.0000$ & $0.00\%$ & \textit{a person slowly jumped forward} \\ 
$0.0110$ & $9.89\%$ & \textit{a man walks counterclockwise in a circle} \\
$0.0283$ & $22.09\%$ & \textit{the person quickly walks forward, and picks something up.} \\
$0.0361$ & $31.49\%$ & \textit{a person picks an item up and moves it a foot to their right and places it down.} \\
$0.0442$ & $44.96\%$ & \textit{a man sidesteps suddenly to his left, bumps into something and leans over, looks around, then walks to his left, bumping into something else and once more leaning over.} \\
$0.0502$ & $55.07\%$ & \textit{someone walks with difficulty on their right side, then tries to run} \\
$0.0616$ & $71.70\%$ & \textit{a person stretches their hips, then arms, then bends forwards and steps forwards.} \\
$0.0784$ & $86.40\%$ & \textit{a person takes in big steps in a hurry walking into the rectangular area while hands are dangling and swinging.} \\
$0.0866$ & $90.87\%$ & \textit{someone puts both of their hands on their chests and appears to be laughing. then waves their left hand.} \\
$0.1051$ & $96.04\%$ & \textit{a person crosses his arms in an x-shape out in front of him and then quickly swings them to the side, brushes off is left leg with his left hand, and then raises his left hand as if to wave.} \\
$0.1872$ & $99.78\%$ & \textit{a person makes several hand gestures and appears to move objects around.} \\
\hline

\end{tabular}}
\end{table*}

\subsection{Analysis on More Diverse Generation}

\paragraph{Metrics on Diverse Generation.} To fairly compare the generalization ability of our proposed \name and other existing works, \eg MotionDiffuse~\cite{zhang2022motiondiffuse}, we propose several new metrics. Specifically, inspired by imbalanced regression task~\cite{ren2022balanced}, here we propose two variants of the original Multimodality Distance. First, we give the definition of sample's \textbf{Rareness}. As for a test prompt $\mathrm{p}$, we calculate its rareness $r_p$ as:
\begin{equation}
    r_p = 1 - \max\limits_i \{<E_T(t_i), E_T(\mathrm{prompt})>\},
\end{equation}
where $E_T$ denotes the text encoder in the CLIP~\cite{radford2021learning} model, $t_i$ is the motion description in the training set, and $<\cdot, \cdot>$ represents the cosine similarity of the two given vectors. Intuitively, this formulation measures the maximum similarity between the given prompt and training prompts. If this similarity is larger, then the rareness will be lower, and vice versa.

Based on the definition of rareness, we sort all samples in increasing order and define the following metrics: \textbf{1) tail 5\% MM}, the average Multimodality Distance of the last 5\% samples; \textbf{2)balanced MM}. we evenly divide the distance space into 100 bins and then calculate the average distance for each bin. Then balanced MM Dist denotes the average distance of all bins. Figure ~\ref{fig:dist} shows the distribution of rareness. The minimum value is almost 0, meaning some captions in the test split are similar to some of the training split. The maximum value is less than 0.25. We divide the whole distribution into 100 bins as the requirement of our proposed \textit{balanced MM}. Most test data concentrate in interval $[0.03, 0.07]$.

In addition, we provide some examples of different rarenesses in Table ~\ref{tab:example}. From these examples, we can find that the increase of rareness usually means the complication of caption in three aspects: unseen expression, more thorough description, and action combination. Some words or phrases are uncommon in the training set, such as `x-shape' and `dangling and swinging'. These sentences may contain unseen motions or are hard to understand by the text encoders. An example of an action combination is that 'a person stretches their hips, arms, then bend forwards and steps forwards' contains four unit actions: `stretch hip', `stretch arm', `bend forward', and `step forward'. The generative models are supposed to act them in a row, which is very challenging to current methods. Hence, these examples build up a more difficult and realistic environment for method evaluation.

\paragraph{Results and Analysis.} Table ~\ref{tab:few_shot} shows the generalization ability of three different methods. As for the baseline model, we simply drop out the retrieval technique. From this table we can find that, with our proposed retrieval technique, \name outperforms both the baseline model and state-of-the-art methods by a remarkable margin.

\subsection{Qualitative Results}
\label{sec:qualitative}
To illustrate the effectiveness of \name, we provide a qualitative comparison between previous works and \name. More examples are available in the project page. As shown in Figure ~\ref{fig:comparison}, \name stands out as the only approach that effectively conveys text descriptions that involve both action and path information. In contrast, Guo \etal's method falls short in capturing path descriptions. MotionDiffuse performs well in action categories, but it lacks precision in providing path details. Meanwhile, MDM captures path information, but its generated actions are incorrect. In the examples evaluated, \name demonstrates its capability to appropriately structure and present the content.

\begin{table}[t]
\centering
\small
\caption{\textbf{Evaluation of Generalization Ability.} All results are reported on the KIT testset. The best results are in \textbf{bold}.}
\label{tab:few_shot}
\setlength{\tabcolsep}{1.4mm}
{
\begin{tabular}{cccc}
\hline\textbf{}

Method & MM $\downarrow$ & tail 5\% MM $\downarrow$ & balanced MM$\downarrow$ \\
\hline
MotionDiffuse & 2.958 & 5.928 & 4.285 \\
\hline
Baseline & 3.371 & 6.173 & 4.661 \\
Ours & \textbf{2.814} & \textbf{5.439} & \textbf{4.028} \\
$\Delta$ & 0.557 & 0.734 & 0.633 \\
\hline
\end{tabular}}
\vspace{-10pt}
\end{table}





\section{Conclusion}

In this paper, we present \name, a retrieval-augmented motion diffusion model for text-driven motion generation. Equipped with a multi-modality retrieval technique, the semantics-modulated attention mechanism, and a learnable condition mixture strategy, \name efficiently explores and utilizes appropriate knowledge from an auxiliary database to refine the denoising process without expensive computation. Quantitative and qualitative experiments are conducted to demonstrate that \name has achieved superior performance in text-driven motion generation, particularly for uncommon motions. 

\noindent\textbf{Social Impacts.} This technique can be used to create fake media when combined with 3D avatar generation. The manipulated media conveys incidents that never truly happened and can serve malicious purposes.



\clearpage
{\small
\bibliographystyle{ieee_fullname}
\bibliography{arxiv_paper}
}

\end{document}